\documentclass[sigconf,authorversion,nonacm]{acmart}
\usepackage{float}
\usepackage{multirow}
\usepackage{textcmds}
\newcommand{\etal}{{\textit{et al.}}}
\AtBeginDocument{%
  \providecommand\BibTeX{{%
    Bib\TeX}}}

\begin{document}

\title{IIITD-20K: Dense captioning for Text-Image ReID}



\author{A V Subramanyam}
\affiliation{%
  \institution{IIITD, India}
  \country{}
}
\email{subramanyam@iiitd.ac.in}
\author{Niranjan Sundararajan}
\authornotemark[1]
\affiliation{%
  \institution{IIITD, India}
  \country{}
  }
\email{niranjan20090@iiitd.ac.in}
\author{Vibhu Dubey}
\authornotemark[1]
\affiliation{%
  \institution{IIITD, India}
  \country{}
}
\email{vibhu20150@iiitd.ac.in}
\author{Brejesh Lall}
\affiliation{%
 \institution{IITD, India}
 \country{}
 }
 \email{brejesh@ee.iitd.ac.in}

\begin{abstract}
  Text-to-Image (T2I) ReID has attracted a lot of attention in the recent past. CUHK-PEDES, RSTPReid and ICFG-PEDES are the three available benchmarks to evaluate T2I ReID methods. RSTPReid and ICFG-PEDES comprise of identities from MSMT17 but due to limited number of unique persons, the diversity is limited. On the other hand, CUHK-PEDES comprises of 13,003 identities but has relatively shorter text description on average. Further, these datasets are captured in a restricted environment with limited number of cameras. In order to further diversify the identities and provide dense captions, we propose a novel dataset called IIITD-20K. IIITD-20K comprises of 20,000 unique identities captured in the wild and provides a rich dataset for text-to-image ReID. With a minimum of 26 words for a description, each image is densely captioned. We further synthetically generate images and fine-grained captions using Stable-diffusion and BLIP models trained on our dataset. We perform elaborate experiments using state-of-art text-to-image ReID models and vision-language pre-trained models and present a comprehensive analysis of the dataset. Our experiments also reveal that synthetically generated data leads to a substantial performance improvement in both same dataset as well as cross dataset settings. Our dataset is available at https://bit.ly/3pkA3Rj.
\end{abstract}


\maketitle
\def\thefootnote{*}\footnotetext{Equal contribution}
\section{Introduction}
Text-to-image (T2I) ReID involves matching a query text description of a person against a set of gallery images to retrieve the target identity \cite{li2017person}. This cross-modal task is extremely challenging compared to uni-modal image ReID where both query and gallery are images. The text descriptions are often generic and coarse-grained in nature, whereas, images carry dense fine-grained visual information. Consequently, the given textual description can correspond to different images with different visual details, leading to large inter-class variance in textual features. Thus, T2I ReID suffers from large modality gap and is much more challenging than image-based ReID. Additionally, because textual descriptions are often ambiguous and subjective, it is difficult to accurately match them with visual features. Therefore, developing effective methods for T2I ReID is an important and challenging research direction in the field of computer vision and NLP.

CUHK-PEDES \cite{li2017person} was the first attempt to build a T2I ReID dataset. The dataset was created from 5 different datasets, CUHK03 \cite{li2014deepreid}, Market-1501 \cite{zheng2015scalable}, SSM \cite{xiao2016end}, VIPER \cite{gray2007evaluating} and CUHK01 \cite{li2013human}. Each image is supplied with two captions with an average word length of 23.5. Some captions also capture actions and background details \cite{ding2021semantically}. Images of CUHK-PEDES are obtained under similar conditions and to address these limitations, RSTPReid \cite{zhu2021dssl} and ICFG-PEDES \cite{ding2021semantically} use images from MSMT17 which are captured under complex conditions compared to that of CUHK-PEDES. RSTPReid has two captions per image with a minimum word length of 23. Further, identifying the shorter average word-length and irrelevant details of captions in CUHK-PEDES, ICFG-PEDES was proposed by Ding \etal. It has lesser number of identities but the average word length is about 35. Additionally, the descriptions are identity centric and avoid action or background details. 

In this paper, we propose a new dataset for text-to-image ReID. Our dataset has an average word length of 36 which is 1.56$\times$ that of CUHK-PEDES. We also have 4.87$\times$ and 1.53$\times$ identities compared to RSTPReid, ICFG-PEDES and CUHK-PEDES, respectively. Images in our dataset are scraped from publicly available sources and thus do not have any environment restrictions, unlike the existing datasets. Thus, it provides a large and diverse dataset to accelerate research in the field of text-to-image ReID.

\begin{table*}[ht]
\centering
\caption{Datasets}
\begin{tabular}{l|l|l|l|l|l|l|l|l} 
\hline
Dataset  & Avg word &unique   & Unique  & Min. words/ & Max. words/ & images & Captions/ & Cameras\\ 
& length & IDs &words &caption & caption& &image & \\
\hline
CUHK-PEDES & 23.5& 13,003& 9,408 & 12 & 96 & 40,206 &2 & \multirow{ 3}{*}{$\ll$ IDs} \\ 
RSTPReid & 25.75 &4,101& 4,628 &11 &68 &20,505 & 2&\\ 
ICFG-PEDES & 34.86 & 4,102& 4,411 &9 & 79&54,522 & 1& \\ 
\hline
IIITD-20K & 35.9 & 20,000 & 5,281& 26 & 97 & 20,000 & 2 & $\approx$ IDs\\ 
\hline
\end{tabular}
\label{tab:dataset-compare}
\end{table*}

\section{Related Work}
Cross modal text based object recognition has witnessed a surge in the interest in the recent past. This is primarily driven by the efficient deep learning models developed in vision and NLP community \cite{cao2022image}. These methods can broadly be categorised into Global Feature Embedding Based Methods and Attention Based Re-ID Methods.

\subsection{Global Feature Embedding}
Some of the early deep learning based methods focused on extracting global features \cite{frome2013devise, faghri2017vse++} and applied various loss functions such as N-pair \cite{sohn2016improved} and bi-rank \cite{liu2017learning}.
Zhang \etal ~\cite{zhang2018deep} introduced cross-modal loss functions for projection matching (CMPM) and classification (CMPC) to learn discriminative text-image representations. However, global feature embedding based strategy ignores the importance of fine-grained visual semantic similarities of text-image pairs. As a consequence, the learned representations suffer from irrelevant details such as  background or action. To address this, attention based methods are proposed which we discuss next.

\subsection{Attention based models}
 Li \etal ~\cite{li2017person} propose a GNA-RNN approach which uses a combination of visual sub-network and language sub-network to effectively construct word-image relations and encode both language and visual information. The unit-level attention and word-level gates weigh the visual units according to the input word and importance of different words, respectively. Further, the aggregation of all unit activations generates the final affinity between the sentence and the person image. Attention based mechanism has also been applied in other works \cite{li2017identity, huang2017instance, chen2018improving, liu2019deep, zheng2020dual, ding2021semantically, zhu2021dssl, aggarwal2020text}. 

 With the phenomenal performance of transformers, NLP models such as BERT \cite{devlin2018bert} have been employed in HGAN \cite{zheng2020hierarchical}, NAFS \cite{gao2021contextual}, LapsCore \cite{wu2021lapscore}, LGUR \cite{shao2022learning}.

\subsection{Vision-Language models} Inspired by the success of BERT-like models, vision-language has attracted a great deal of attention. The vision-language models aim to learn joint representations of images and their corresponding textual descriptions, which can be applied to downstream tasks like image captioning, retrieval, visual question answering, among others. These models can be broadly classified into two-stream models and single-stream models. Two-stream models, such as ViLBERT \cite{lu2019vilbert} and LXMERT \cite{tan2019lxmert}, process text and images separately using two modules. In contrast, single-stream models, such as VisualBERT \cite{li2019visualbert}, VL-BERT, and UNITER, encode both modalities within the same module. The vision-language pretrained models such as CLIP \cite{radford2021learning}, ALIGN \cite{jia2021scaling} and BLIP \cite{li2022blip} have shown tremendous improvement in text based retrieval tasks. IRRA \cite{jiang2023cross} shows tremendous improvement in text based person ReID.

\subsection{Baselines} 
We chose three diverse baselines, namely IRRA \cite{jiang2023cross}, LGUR \cite{shao2022learning}  and SSAN \cite{ding2021semantically}, to analyze our proposed dataset.

IRRA \cite{jiang2023cross} uses a cross-modal approach to learn the mapping of visual and textual modalities into a common latent space, allowing for effective matching of multimodal data. It uses a novel similarity distribution matching (SDM) loss which minimizes the KL-divergence between image-text similarities and the distribution obtained from the labels.

 LGUR \cite{shao2022learning} comprises of two modules: Dictionary-based Granularity Alignment (DGA)
and Prototype-based Granularity Unification (PGU). DGA bridges the gap between image and text modality by reconstructing both visual and textual features using a transformer based multi-modality shared dictionary. As DGA is based on reconstruction task, without an explicit guidance it does not learn robust discriminative features. To address this issue, PGU combines the two modalities by projecting them into a shared feature space using a set of shared and learnable prototypes. The shared prototypes have an added benefit that it significantly reduces the computational cost of LGUR compared to methods using cross-modal attention operations.

Ding \etal ~introduce a new model called Semantically Self-Aligned Network (SSAN) \cite{ding2021semantically}. SSAN utilizes contextual cues in language descriptions to extract part-level visual and textual features by inferring word-part correspondences. However, such mapping disregards correlations between body parts and spatial relationships between image regions specified in textual descriptions. Therefore, SSAN  proposes a Multi-View Non-Local Network (MV-NLN) to capture the relationships between body parts. Further, to overcome the intra-class variance in descriptions, the authors propose a Compound Ranking (CR) loss that includes both strong and weak supervision components and acts as a novel data augmentation strategy.

\section{Proposed Dataset}
\begin{figure}
\centering
\includegraphics[width = .5\textwidth]{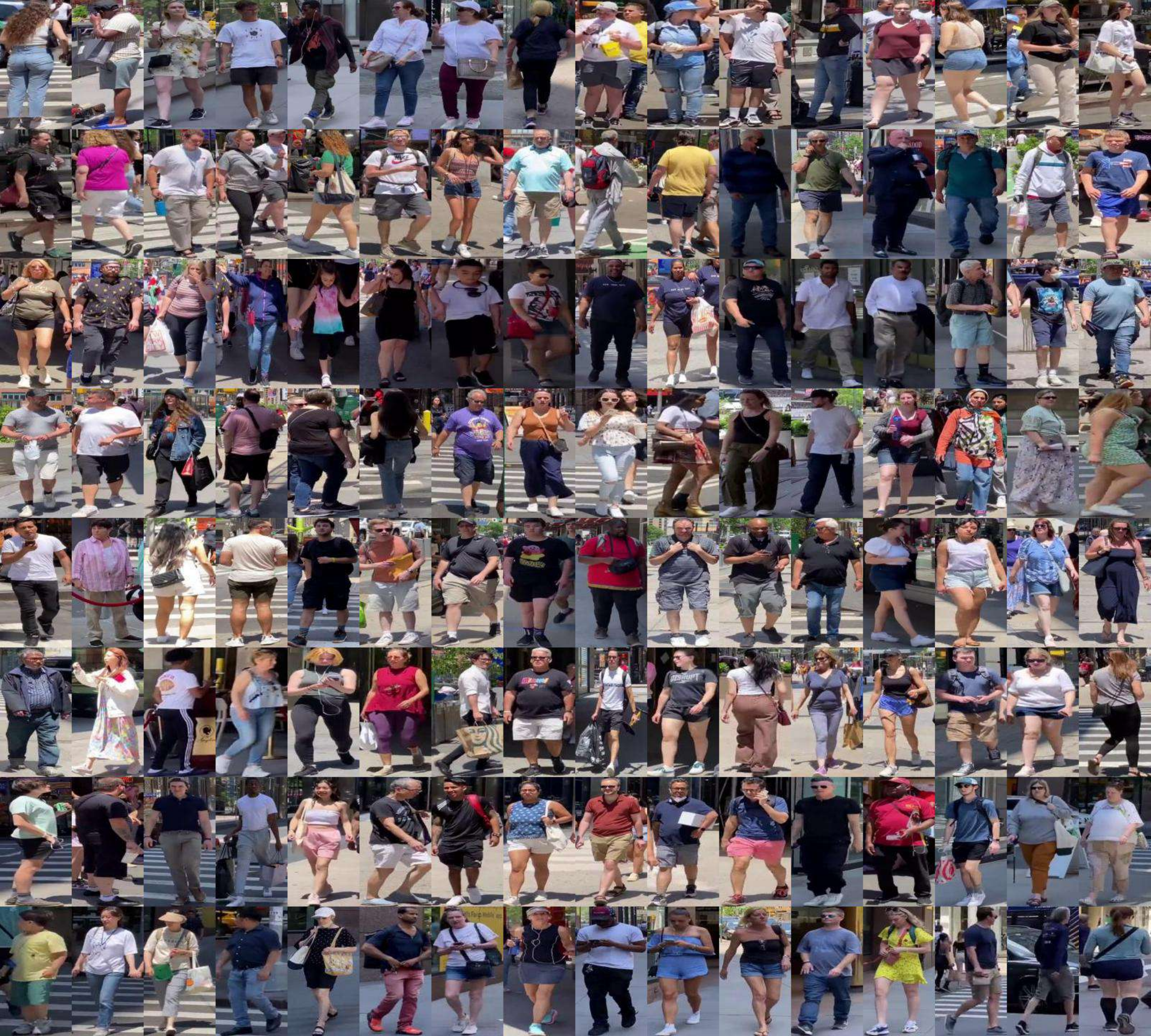}
\caption{Sample images from the dataset.}
\label{fig:sample}
\end{figure}  

Our dataset is collected with the objective of having a large number of unique identities captured under unrestricted environment and captioned by multiple annotators. The images are scraped from publicly available sources in web and tightly cropped to contain full human body. To scrape the images, we use the following keywords, 
\begin{enumerate}
    \item City names: Berlin, London, Newyork, Bangalore, Chennai, Paris, Kolkata, Sydney, Brisbane, Tokyo, California, Delhi, Mumbai;
    \item Years of upload: 2012-2022
\end{enumerate}

\begin{figure}
\centering
\includegraphics[width = .5\textwidth]{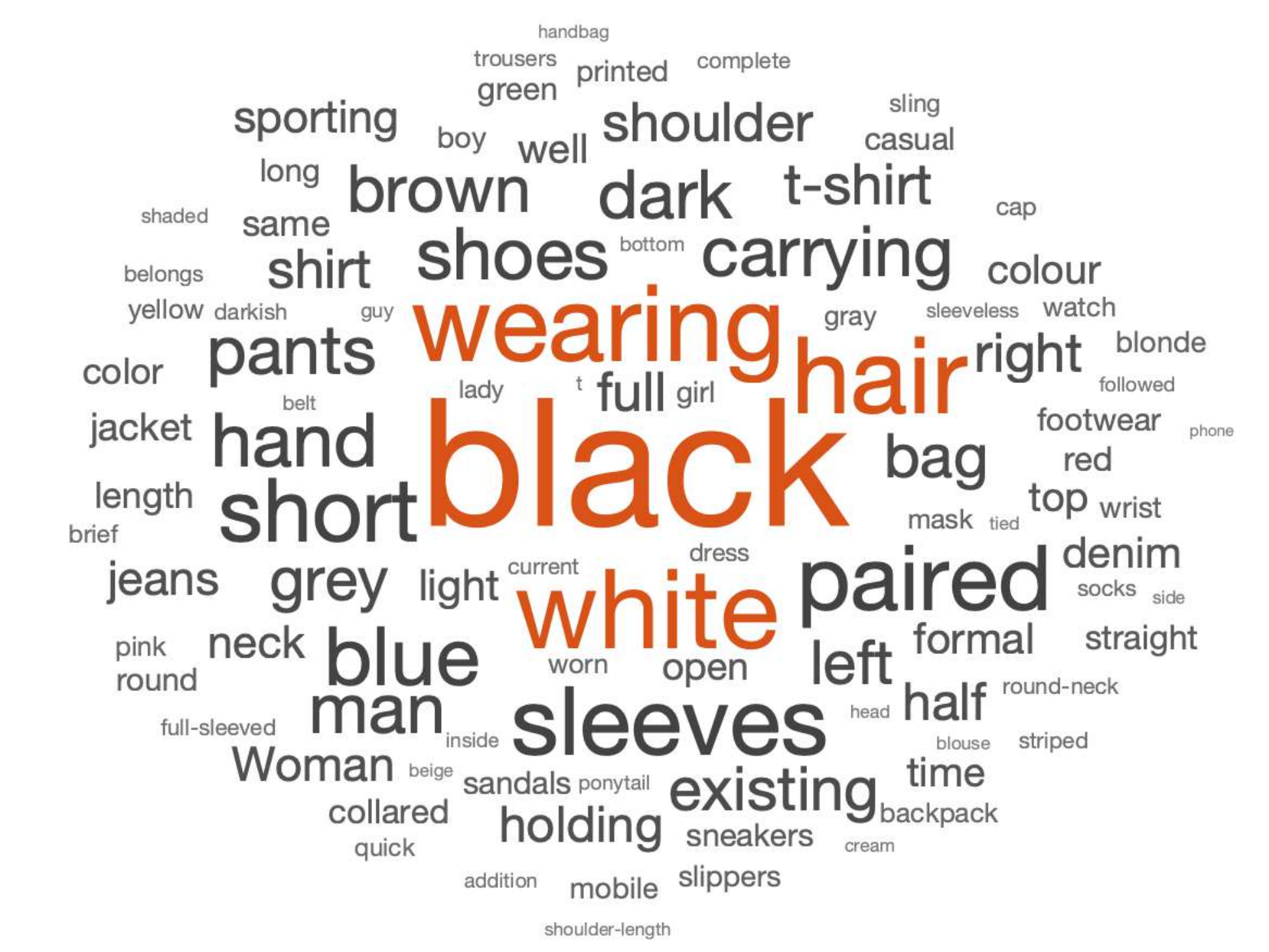}
\caption{High frequency words in the proposed dataset.}
\label{fig:Wordcloud}
\end{figure}

Our dataset has 22,727 images out of which 20,000 are captioned with each image having two captions. There are 14,72,005 words and 53,358 sentences. A total of 30,874 captions have words in the range of 20-40, and, 8,889 captions have words in the range of 40-60. Poor quality images with noise or occlusion were discarded. We show some sample images in Figure \ref{fig:sample}. The wordcloud in Figure \ref{fig:Wordcloud} depicts the high frequency words.

\begin{figure}
\centering
\includegraphics[width = .5\textwidth]{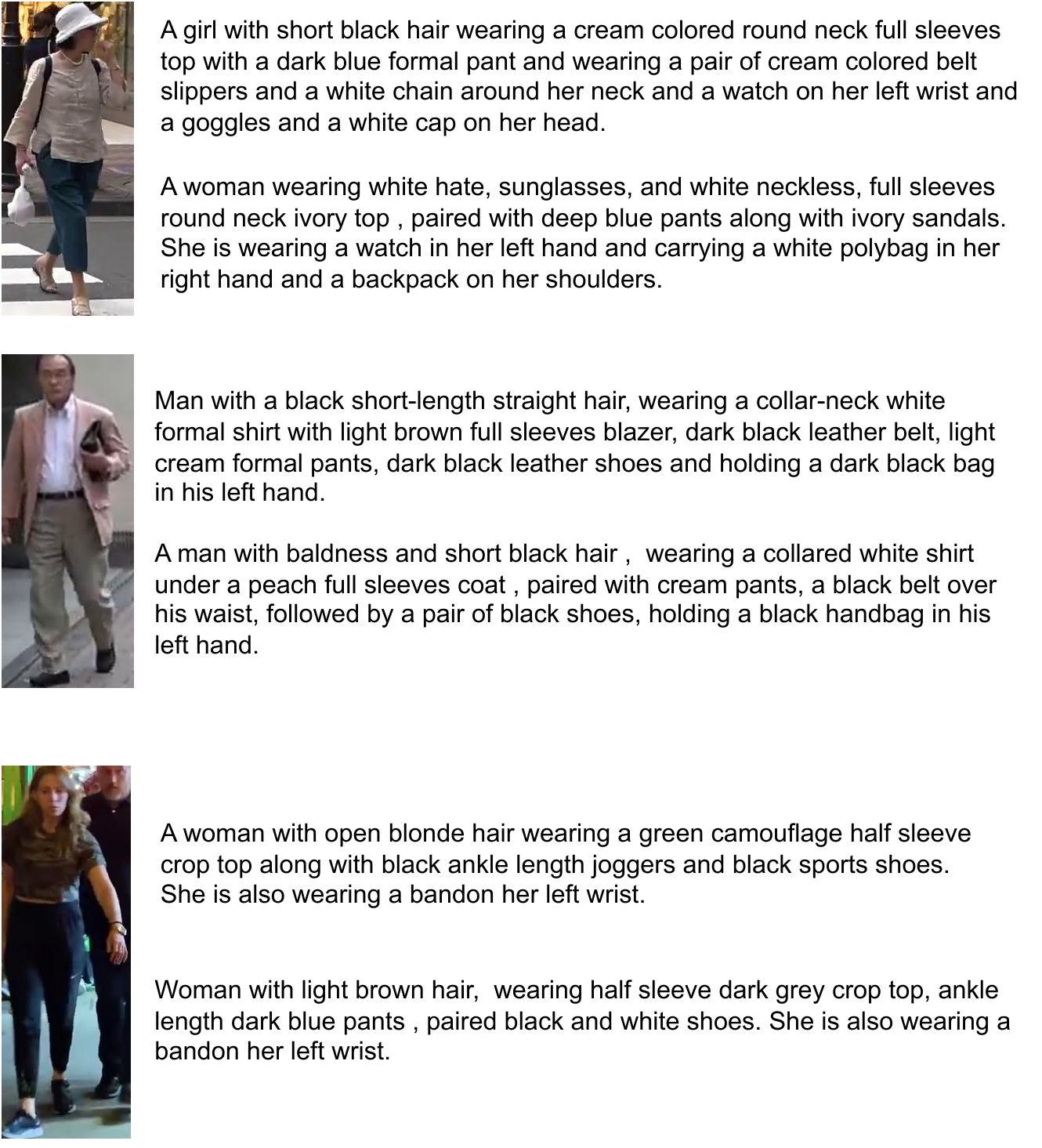}
\caption{Sample images and captions from proposed dataset.}
\label{fig:captions}
\end{figure}

In Figure \ref{fig:captions}, we show the captions corresponding to the images. For the first image, we can see that the annotation captures all the attributes - white cap, goggles, chain, watch and the color of dress. While the first caption does not capture the bag in the right hand, the second annotation takes care of it. Similarly, the other captions also elaborately explain the appearance.

We compare all the four datasets in terms of unique words and IDs, number of images, minimum and maximum number of words per captions and number of captions per image in Table \ref{tab:dataset-compare}. To obtain this we follow the tokenization given in \cite{li2017person} and process all four datasets using the same method. We observe that our dataset scores better in terms of unique IDs and words, and also in average, minimum and maximum number of words per caption. We would like to mention that the number of unique words and average words per caption given in ICFG-PEDES \cite{ding2021semantically} and RSTPReid \cite{zhu2021dssl} differ from the numbers given in Table \ref{tab:dataset-compare}. This may be due to different tokenizations.

\subsection{Synthetic Dataset}
\label{sec:synthetic}
Data augmentation in the form of synthetic dataset can be useful in improving the performance. Generative models such as latent diffusion models (LDM) \cite{rombach2022high} and BLIP \cite{li2022blip} have shown tremendous progress in high definition image synthesis and caption generation, respectively. In order to leverage these commanding models, we first finetune LDM with our dataset. The latent diffusion model is a conditional UNet model \cite{ronneberger2015u} and is trained to predict the noise in the given input image. The text conditioning is based on a cross-attention model \cite{vaswani2017attention}. In our case, we use the image-text pairs as input. As the finetunning of LDM suffers from catastrophic forgetting, we adopt the LoRA training technique proposed in \cite{hu2021lora}. LoRA not only overcomes catastrophic forgetting, it has lesser number of parameters to be trained as well as a high learning rate can be used. At the inference, we use the training captions as prompt and sample images from the finetuned LDM model using DDPM sampling. 

On the other hand, BLIP is a powerful captioning model which is trained using a large corpus and employs nucleus sampling \cite{holtzman2019curious} to generate diverse captions. However, as it is trained using datasets which only have salient points as captions, the generated captions lack fine-grained nature needed for image-text ReID. Therefore, to generate captions which are more detailed and capture different attributes of the image, we finetune the BLIP model using our dataset. 

The images generated from LDM are then fed to the finetuned BLIP model to obtain the respective captions. The images generated from LDM as well as captions generated from BLIP now act as image-caption pairs. We further use this synthetic data and train the ReID models. In our experiments, we find that augmenting the training with this synthetic data gives a signifcant boost.

\section{Evaluation Protocol}
Here we discuss the protocol for evaluation. We divide the dataset into train, validation and test split, following the practice in CUHK-PEDES. The train set has 15K image-text pairs and, validation and test set have 2.5K image-text pairs each. There is no overlap between train, validation or test sets. As the dataset has 22,727 images in total, for the remaining 2,727 images, the captions were rejected. We do not include these images in the evaluation, however, they can be included as distractors. 

\section{Implementation Details}
In case of BLIP, we use ViT-B as image encoder \cite{dosovitskiy2020image} and BERT$_{\text{base}}$ \cite{dosovitskiy2020image} as text encoder. These models are initialized with the checkpoints provided in \cite{li2022blip}. We use Adam with a learning rate of 1e-6, weight decay 5e-2 and train for only one epoch. We keep the other settings similar to the finetuning process given in \cite{li2022blip}.

The LDM \cite{rombach2022high} uses Hugging Face' \qq{stable diffusion v1-4}\footnote{https://huggingface.co/CompVis/stable-diffusion-v1-4} model and CLIP based text encoder \cite{radford2021learning}. We train the model for 23K steps with a batch size of 12 and a learning rate of 1e-4.

IRRA uses the CLIP model \cite{radford2021learning}. LGUR uses DeiT-Small \cite{touvron2021training} image backbone and BERT \cite{devlin2018bert} as text-encoder. SSAN uses ResNet-50 as image encoder and Bi-LSTM as text encoder. We use the same settings provided in their respective papers and released codes to train the models.

\section{Experiments}
\textbf{CUHK-PEDES} has 40,206 images
and 80,412 textual descriptions for 13,003 identities. We use the train set comprising of 
11,003 identities, the validation and test comprising of 1,000 identities each, respectively. 

\noindent{\textbf{ICFG-PEDES}} has a train set 
of 3,102 identities, and a test set of 1,000 identities. 

\noindent{\textbf{RSTPReid}} contains 4,101 identities. The training set comprises of 3,701 identities, while, validation and test set contain 200 identities each, respectively.

\noindent{\textbf{Evaluation metrics}} We use the widely popular metrics - Rank-k and mean Average Precision (mAP) as the evaluation criteria. Rank-k computes the probability of finding at least one relevant image in the top-k retrieved images. Additionally, mAP computes the average precision for each query and then takes the mean of those average precision values over all queries.

\subsection{Synthetic Images and Captions}
\begin{figure}[!h]
\centering
\includegraphics[width = .5\textwidth]{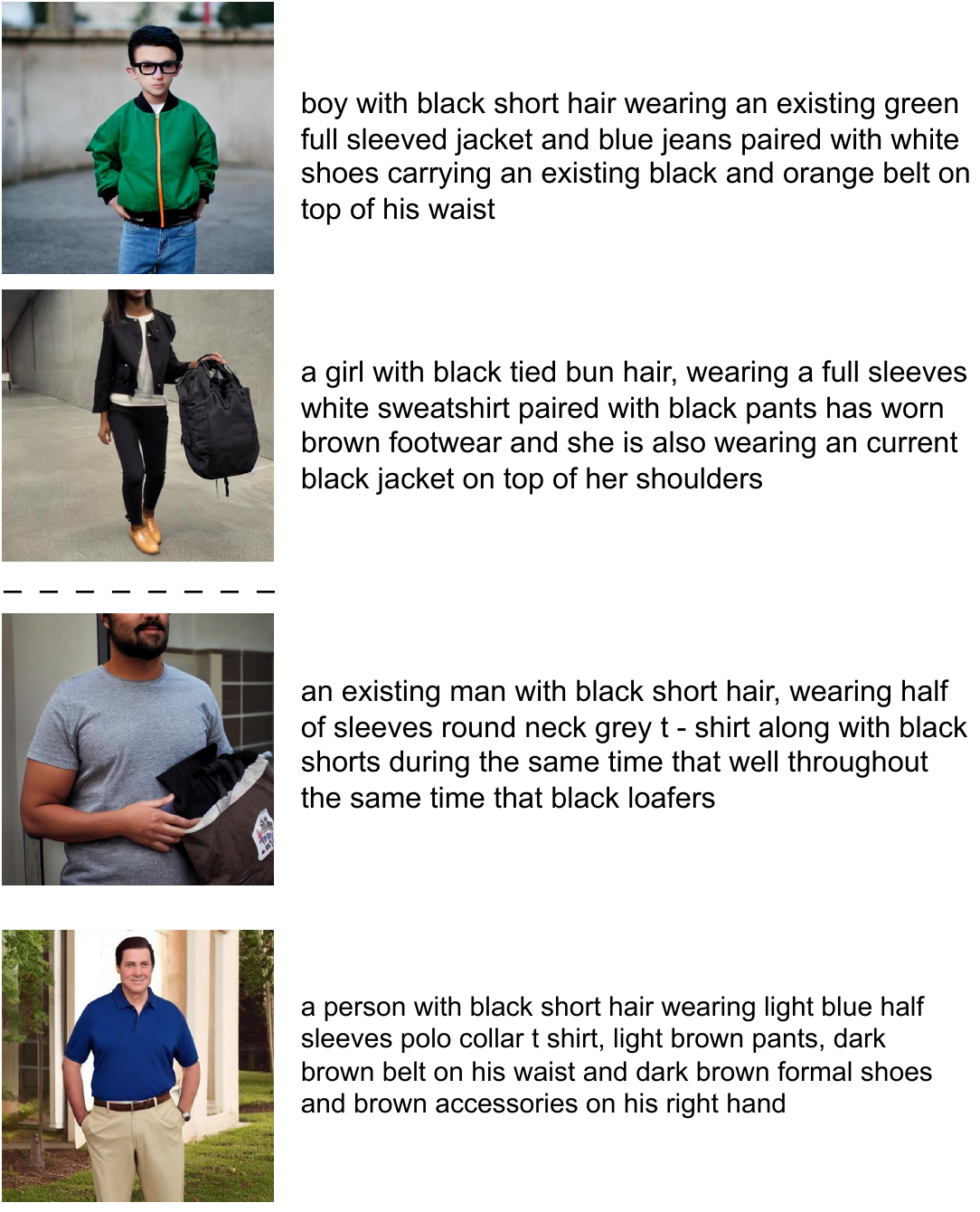}
\caption{Synthetic images and captions. Top two rows: Generated images from LDM \cite{rombach2022high} and captions from BLIP \cite{li2022blip} for the train set of our dataset. Bottom two rows: generated samples for CUHK-PEDES.}
\label{fig:synthetic-captions}
\end{figure}

The synthetic data is generated as given in Section \ref{sec:synthetic}. We generate two images per person ID, and one caption per image. In total, we generate 30K more images and captions. We assign the same labels as that of original dataset. In Figure \ref{fig:synthetic-captions}, we show the images generated from LDM and respective captions generated from finetuned BLIP model. We see that the caption corresponding to top most image is partially correct. The attributes like \qq{green jacket} and \qq{blue jeans} match with the image. Whereas, \qq{white shoes} and \qq{belt} part is missing in the image. Similarly, in the second image, the \qq{bun hair} part does not match with the image. Even though these descriptions are noisy, they are helpful in training the model.

Additionally, we generate 15K captions from BLIP model using original images of our dataset. We use these captions along with original images of our dataset during training to analyze the goodness of these captions.

\subsection{Ablation Study}
\subsection*{Proposed dataset}
Here, we analyse the results obtained from synthetic data in Table \ref{tab:synthetic}. We analyse our dataset using state-of-art baselines IRRA \cite{jiang2023cross}, LGUR \cite{shao2022learning} and SSAN \cite{ding2021semantically}. We first compute the results for same dataset setting using LGUR and IRRA. In Table \ref{tab:synthetic}, \qq{Ours} denote that original dataset is used for training. B1 and B2 denote batches, where B1 indicates one set of LDM-BLIP generated 15K image-text pairs and B2 denotes another set of 15K image-text pairs. We report the results for various combinations of original dataset and synthetic dataset used for training. 

LGUR gives an R-1 of 65.96\% and IRRA obtains 77.08\% when trained using original dataset, that is the variant \qq{Ours} in Table \ref{tab:synthetic}.
If we only use original images and BLIP captions, we observe an R-1 of 17.94\%. This indicates that though the captions are noisy, they may help in augmenting the original captions.

\subsubsection*{BLIP} We now discuss the results for the case where we use original dataset (both images and captions) along with captions generated from BLIP. As BLIP captions are noisy, we assign a lower weight to the loss computed for the BLIP captions. We denote this weight by $w$. We see that the variant \qq{Ours+BLIP ($w$=0.1)} gives better performance for LGUR. Here, $w=0.1$ means that the loss computed using synthetic captions is given a weight of 0.1 which is 10$\times$ lower than the weight given to loss computed using original captions. Similarly, $w=1$ indicates that the losses computed using original dataset and synthetic dataset have same weight. However, this setting gives inferior results compared to $w=0.1$ variant. We also perform annealing learning rate but did not find good performance. 

\begin{table}[!ht]
\centering
\caption{Same dataset performance using synthetic data. B1+B2 indicates that both batches were used during the training}
\begin{tabular}{l|l|l|l} 
\hline
Model&Setting    & R-1 & mAP  \\ 
\hline
\multirow{ 11}{*}{LGUR} & Ours & 65.96 & 71.06\\
& BLIP & 17.94 & 23.48 \\
&Ours+BLIP ($w$=0.1)   & 66.70 & 72.02\\ 
&Ours+BLIP ($w$=1)  & 64.66 & 70.27\\ 
&Ours+BLIP (anneal) & 64.82 & 70.18\\
\cline{2-4}
& B1 BLIP+LDM &15.46 & 20.16\\
& B1+B2 BLIP+LDM & 21.32 & 26.80 \\
&Ours+B1+B2  & 67.10 & 72.18\\
& BLIP+LDM ($w$=0.1) & &\\
&Ours+B1+B2  & \textbf{67.12} & \textbf{72.46}\\
& BLIP+LDM ($w$=1) & & \\
\hline
\multirow{ 9}{*}{IRRA} & Ours & 77.08 & 83.10  \\
& Ours+BLIP ($w$=0.1)  & 78.2 &83.98 \\
&Ours+BLIP ($w$=1)  &  76.26 & 82.34\\
\cline{2-4}
&B1 BLIP+LDM & 35.44 & 45.14\\
& B1+B2 BLIP+LDM & 37.06 & 46.99\\
&Ours+B1+B2 & 77.54 & 83.48\\
& BLIP+LDM ($w$=0.1) & &\\
&Ours+B1+B2  &\textbf{78.94} & \textbf{84.56}\\
& BLIP+LDM ($w$=1) & &\\
\hline
\end{tabular}
\label{tab:synthetic}
\end{table}

\subsubsection*{BLIP+LDM} On the other hand, when we use only synthetic data (B1 BLIP + LDM), that is images from LDM and captions from BLIP, we get an R-1 of 15.46\% for LGUR. R-1 increases to 21.32\% when we add more synthetic data (B1+B2 BLIP+LDM). When we use
 synthetic images and captions along with original dataset, \qq{Ours+B1+B2 BLIP+LDM ($w$=1)} variant gives best performance amongst all settings. LGUR has an R-1 of 67.12\% which is a boost of 1.16\% compared to the case when no synthetic data is used. A similar rise is observed in mAP also. In case of IRRA, $w=1$ variant scores 78.94\% which is a boost of 1.86\% over the non-synthetic case.

\subsubsection*{Synthetic Cross-dataset} In case of cross-dataset setting also in Table \ref{tab:cross-dataset-synthetic}, we see that the best performance is obtained when both synthetic images and captions are used. In case of CUHK-PEDES, the best performance is 36.68\% for LGUR, which is a boost of 2.86\%. We also see an increase by 1.66\% in ICFG-PEDES. However, there is a marginal increase in case of RSTPReid. IRRA shows an increase of 2.97\% in CUHK-PEDES and has a substantial boost in ICFG-PEDES.

\begin{table}[h]
\centering
\caption{Cross-dataset performance using synthetic data. Rank-1 accuracy is reported }
\begin{tabular}{l|l|l|l|l} 
\hline
Model&Setting    & CUHK & ICFG & RSTP  \\ 
\hline
\multirow{ 8}{*}{LGUR}&Ours & 33.82&	28.91	&31.15\\
&Ours+BLIP ($w$=0.1)   & 34.97	& 29.14	&33.15\\ 
&Ours+BLIP ($w$=1)  & 35.02&	29.02	&32.30\\ 
&Ours+BLIP (anneal) & 34.19	&27.72&	30.40\\
\cline{2-5}
&{Ours+B1+B2 } & 36.39&29.85&	\textbf{33.60}\\
&BLIP+LDM ($w$=0.1) & & &\\
 &Ours+B1+B2 & \textbf{36.68}&	\textbf{30.57}	&31.20\\
 &BLIP+LDM ($w$=1) & & &\\
\hline
\multirow{ 3}{*}{IRRA} & Ours & 43.74	&35.61 &	\textbf{38.00}\\
& Ours+B1+B2  &\textbf{46.71} & \textbf{36.64} & 37.00\\
& BLIP+LDM ($w$=1) & &\\
\hline
\end{tabular}
\label{tab:cross-dataset-synthetic}
\end{table}

\subsection*{CUHK-PEDES}
We also generate 11,003 training image-text pairs synthetically for CUHK-PEDES. Here, we observe an increment of 0.44\%. As both LDM and BLIP are trained on our dataset, the generated pairs are more noisier for CUHK-PEDES and the increment is not as much as in our dataset. However, if both LDM and BLIP are trained using CUHK-PEDES, we believe that there will be substantial boost in these results.
\begin{table}[!ht]
\centering
\caption{Performance on CUHK-PEDES using synthetic data. Rank-1 accuracy is reported}
\begin{tabular}{l|l} 
\hline
Setting& R-1   \\ 
\hline
CUHK & 65.35 \\
CUHK+ B1+B2 BLIP+LDM ($w$=0.1) & 65.58\\
CUHK+ B1+B2 BLIP+LDM ($w$=1) & \textbf{65.77}\\
\hline
\end{tabular}
\label{tab:cuhk-synthetic}
\end{table}

\subsection{Performance on same dataset setting}
 In Table \ref{tab:Performance-in-IIITD}, we show the results on all original datasets. Here, the models are trained and tested on same dataset. For our dataset, IRRA achieves 77.08\% R-1 accuracy, while LGUR obtains R-1 accuracy of 65.96\%. SSAN achieves an accuracy of 57.56\%. IRRA achieves the best scores.

\begin{table}[h]
\centering
\caption{Performance comparison on all datasets. Rank-1 accuracy is reported}
\begin{tabular}{l|l|l|l|l} 
\hline
Model  & {CUHK-PEDES} &{ICFG-PEDES}  & {RSTPReid} & {\textbf{Ours}} \\ 
\hline

IRRA & \textbf{73.68}  & \textbf{63.43} & \textbf{57.00}  & \textbf{77.08} \\
LGUR & 65.35  & 59.53 & 48.65 & 65.96 \\ 
SSAN & 60.71 & 54.05 & 38.80 & 57.56 \\
\hline
\end{tabular}
\label{tab:Performance-in-IIITD}
\end{table}

\begin{table*}[ht]
\caption{Cross-dataset T2I: Trained on one dataset and test on remaining test sets} 
\centering
{
\begin{tabular}{c c c| c c| c c | c c| c c |c c}
\hline
 Train &\multicolumn{6}{c|}{CUHK-PEDES} &\multicolumn{6}{c}{ICFG-PEDES}  \\
 \cline{1-13}
  Test&\multicolumn{2}{c|}{ICFG-PEDES} &\multicolumn{2}{c|}{RSTPReid} & \multicolumn{2}{c|}{\textbf{Ours}} 
  &\multicolumn{2}{c|}{CUHK-PEDES} &\multicolumn{2}{c|}{RSTPReid} & \multicolumn{2}{c}{\textbf{Ours}} \\
  \cline{1-13}
& R-1 & mAP& R-1 & mAP& R-1 & mAP& R-1 & mAP &R-1 & mAP& R-1  &mAP\\
\hline 
IRRA & 43.13 & 23.37 & 53.30 & 40.05 & 64.56 & 72.84 & 33.48 & 31.56 & 45.30 & 36.82 & 43.82 & 53.79 \\
 LGUR & 34.52 & 15.85 & 41.30 & 27.98 & 54.42 & 60.49 & 26.54 & 21.63 & 46.25&33.09 & 28.82&34.81 \\
SSAN & 24.47 & 10.59 & 13.75 & 8.935 & 37.12 & 43.83 & 15.50 & 13.37 & 35.50 & 28.30 & 16.36 & 21.38 \\
\hline\\
\hline
Train&\multicolumn{6}{c|}{RSTPReid} & \multicolumn{6}{c}{\textbf{Ours}} \\
 \cline{1-13}
 Test &\multicolumn{2}{c|}{CUHK-PEDES} &\multicolumn{2}{c|}{ICFG-PEDES} & \multicolumn{2}{c|}{\textbf{Ours}} 
  &\multicolumn{2}{c|}{CUHK-PEDES} &\multicolumn{2}{c|}{ICFG-PEDES} & \multicolumn{2}{c}{RSTPReid} \\
 \cline{1-13}
&R-1 & mAP &R-1 & mAP &R-1 & mAP& R-1 & mAP &R-1 & mAP & R-1 & mAP\\
\hline 
 IRRA & 32.49 & 30.30 & 31.23 & 19.36 & 26.32 & 35.79 & 43.74 & 39.72 & 35.62 & 17.51 & 38.00 & 28.57 \\
 LGUR & 15.50&13.06& 26.48&15.71&10.66&14.89 &33.82&27.91 &28.91&12.12 & 31.15&21.17\\
 SSAN & 10.04 & 8.39 & 20.19 & 12.15 & 4.92 & 7.83 & 21.86 & 17.96 & 18.65 & 7.26 & 9.90 & 6.12 \\
\hline
\end{tabular}
}
\label{tab:T2I}
\end{table*}

\begin{table*}[ht]
\caption{Cross-dataset I2T} 
\centering
{
\begin{tabular}{c c c| c c| c c | c c| c c| c c}
\hline
Train &\multicolumn{6}{c|}{CUHK-PEDES} &\multicolumn{6}{c}{ICFG-PEDES}  \\
 \cline{1-13}
Test  &\multicolumn{2}{c|}{ICFG-PEDES} &\multicolumn{2}{c|}{RSTPReid} & \multicolumn{2}{c|}{\textbf{Ours}} 
  &\multicolumn{2}{c|}{CUHK-PEDES} &\multicolumn{2}{c|}{RSTPReid} & \multicolumn{2}{c}{\textbf{Ours}} \\
  \cline{1-13}
& R-1 & mAP& R-1 & mAP& R-1 & mAP& R-1 & mAP &R-1 & mAP& R-1  &mAP\\
\hline 
IRRA & 42.38 & 20.51 & 54.70 & 35.80 & 73.52 & 71.79 & 48.96 & 27.91 & 64.80 & 40.33 & 51.60 & 51.60 \\
 LGUR & 30.71 & 14.55 & 44.9 & 27.65 & 59.84 & 56.7 & 37.54 & 20.22 &60.60 & 36.01 & 35.56 & 33.62 \\
 SSAN & 22.45 & 10.10 & 16.60 & 11.24 & 44.96 & 40.44 & 22.45 & 11.42 & 52.50 & 30.56 & 21.12 & 19.53 \\
\hline\\
\hline
Train&\multicolumn{6}{c|}{RSTPReid} &\multicolumn{6}{c}{\textbf{Ours}} \\
 \cline{1-13}
Test  &\multicolumn{2}{c|}{CUHK-PEDES} &\multicolumn{2}{c|}{ICFG-PEDES} & \multicolumn{2}{c|}{\textbf{Ours}} 
  &\multicolumn{2}{c|}{CUHK-PEDES} &\multicolumn{2}{c|}{ICFG-PEDES} & \multicolumn{2}{c}{RSTPReid} \\
 \cline{1-13}
&R-1 & mAP &R-1 & mAP &R-1 & mAP& R-1 & mAP &R-1 & mAP & R-1 & mAP\\
\hline 
 IRRA & 41.77 & 25.38 & 36.94 & 18.85 & 33.92 & 34.91 & 58.17 & 35.39 & 35.62 & 16.14 & 43.40 & 26.13 \\
 LGUR & 19.91 & 10.96 & 33.61 & 16.11 & 13.12 & 13.08 & 45.41 & 25.06 & 26.30 & 11.88 & 35.60 & 21.72 \\
 SSAN & 13.01 & 6.66 & 24.82 & 11.02 & 7.04 & 6.67 & 28.85 & 15.31 & 16.67 & 6.915 & 11.60 & 8.18 \\
\hline
\end{tabular}
}
\label{tab:I2T}
\end{table*}

\subsection{Cross-dataset performance}
\subsection*{Text-to-Image}
In Table \ref{tab:T2I}, we present the results for cross-dataset testing. Here, we train on one dataset and test on test sets of the remaining three datasets. The evaluation protocol is the same as given in respective datasets. 

\subsubsection*{Train on CUHK-PEDES} We first train on CUHK-PEDES and test on ICFG-PEDES, RSTPReid and ours. IRRA gives best results on all datasets and across all settings. It achieves 43.13\% on ICFG-PEDES, 53.30\% on RSTPReid and 64.56\% on our dataset. In case of LGUR, we obtain an R-1 of 34.52\% on ICFG-PEDES, 41.30\% on RSTPReid, and 54.42\% on our dataset. 

\subsubsection*{Train on Ours} When we train IRRA on our dataset and test on CUHK-PEDES, ICFG-PEDES and RSTPReid, we obtain respective R-1 scores of 43.75\%, 35.62\% and 38.00\%. These are followed by LGUR and SSAN. 

We also infer that the reason behind such strong cross-dataset performance for CUHK-PEDES is the rich vocabulary which comprises of more than 9K unique words. Though a direct comparison is not applicable, we note that training on our dataset and testing on CUHK-PEDES gives good results compared to that of training on ICFG-PEDES or RSTPReid and testing on CUHK-PEDES. For instance, IRRA trained on the proposed dataset gives an R-1 of 43.75\% on CUHK-PEDES. On the other hand, IRRA trained on ICFG-PEDES gives 33.48\% on CUHK-PEDES. 

Further, when we test IRRA on ICFG-PEDES using training set as RSTPReid and Ours, it gives an R-1 of 31.23\% and 35.62\%, respectively. Even though ICFG-PEDES and RSTPReid are derived from the same dataset (MSMT17) itself, training on proposed dataset gives superior performance.

\subsection*{Image-to-Text}
We present I2T results in Table \ref{tab:I2T}. We have similar observations as in case of T2I. IRRA trained on our dataset scores 58.16\%, 35.62\% and 43.40\% on CUHK-PEDES, ICFG-PEDES and RSTPReid, respectively. LGUR obtains an R-1 of 45.41\% on CUHK-PEDES, 26.30\% on ICFG-PEDES, and 35.60\% on RSTPReid. When we test on CUHK-PEDES using training set as ICFG-PEDES and proposed dataset, we find that training on our dataset performs significantly better for all models on both R-1 and mAP. Thus, training on the proposed dataset obtains rich cross-dataset results.

\section{Conclusion}
In this paper, a new text-image ReID dataset is introduced which is rich in terms of the number of identities, the environment of image acquisition, and textual descriptions. We also employ state-of-the-art baselines and perform thorough experiments to analyze the proposed dataset. Furthermore, generative models such as LDM and BLIP are trained to create synthetic text-image pairs, and extensive experiments are conducted in both same and cross dataset settings using original and synthetic data. The results show that the performance of the models can be significantly improved by using the synthetic data in addition to the original data. Overall, the proposed dataset and the experiments provide valuable insights and a benchmark for future research in the field of text-image ReID.

\bibliographystyle{ACM-Reference-Format}
\bibliography{sample-base}

\end{document}